\documentclass{article}
\usepackage[preprint]{neurips_2020}

\usepackage[utf8]{inputenc} 
\usepackage[T1]{fontenc}    
\usepackage{hyperref}       
\usepackage{url}            
\usepackage{booktabs}       
\usepackage{amsfonts}       
\usepackage{nicefrac}       
\usepackage{microtype}      
\usepackage{graphicx}
\usepackage[vlined,ruled,linesnumbered,noend]{algorithm2e}
\usepackage{amsmath}
\usepackage{adjustbox}
\usepackage{enumitem}
\usepackage{wrapfig}
\usepackage{multirow}
\usepackage{array, boldline, makecell}
\usepackage{array}

\def\alg{{DuoLab}\xspace }
\title{Active Learning for Noisy Data Streams Using \\ Weak and Strong Labelers \\ 
}

\author{%
Taraneh Younesian\\
  Delft University of Technology\\
  \texttt{t.younesian@tudelft.nl} \\
   \And
  Dick Epema\\
  Delft University of Technology\\
  \texttt{d.h.j.epema@tudelft.nl}
   \AND
  Lydia Y. Chen\\
  Delft University of Technology\\
  \texttt{y.chen-10@tudelft.nl}
}

\begin{document}

\maketitle

\begin{abstract}
Labeling data correctly is an expensive and challenging task in machine learning, especially for on-line data streams. Deep learning models especially require a large number of clean labeled data that is very difficult to acquire in real-world problems. Choosing useful data samples to label while minimizing the cost of labeling is crucial to maintain efficiency in the training process. When confronted with multiple labelers with different expertise and respective labeling costs, deciding which labeler to choose is nontrivial.

In this paper, we consider a novel weak and strong labeler problem inspired by humans’ natural ability for labeling, in the presence of data streams with noisy labels and constrained by a limited budget.  
We propose an on-line active learning algorithm that consists of four steps: filtering, adding diversity, informative sample selection, and labeler selection. We aim to filter out the suspicious noisy samples and spend the budget on the diverse informative data using strong and weak labelers in a cost-effective manner. We derive a decision function that measures the information gain by combining the informativeness of individual samples and model confidence. 
We evaluate our proposed algorithm on the well-known image classification datasets CIFAR10 and CIFAR100 with up to 60\% noise. Experiments show that by intelligently deciding which labeler to query, our algorithm maintains the same accuracy compared to the case of having only one of the labelers available while spending less of the budget. 
\end{abstract}

\section{Introduction}


Obtaining a labeled dataset for training machine learning models is a time consuming and expensive task. The common practise of curating data  continuously collects both data and the metadata that serves as labels~\cite{hendrycks:NIPs18:glc} from the public domain. This immediately reudces
the cost of acquiring a high volume of labeled data for deep learning models but introduces the challenge of handling data streams with noisy labels~\cite{han:NIPs18:coteaching,wang2018iterative}, i.e., data that is annotated with wrong class labels. Human experts~\cite{Sheng:2008:SIGKDD:Multilabeler} are sought after to correct labels to enhance the robustness of learning models against label noise. Cognitive science~\cite{Yan:JML14:multipleAnnotator} has shown that humans are better in answering binary questions, such as True/False questions, and are less skilled in multiple-choice questions, e.g., identifying one out of 100 classes in CIFAR-100 benchmark~\cite{kriz-cifar10}. 
It is more expensive to use strong labelers who are skilled in directly pointing out true class labels than weak labelers who can only (dis)agree with the provided labels. To cost effectively correct the noisy labels by experts~\cite{Sheng:2008:SIGKDD:Multilabeler,Dy:ICML17:undertainExperts}, it is imperative to assign the correct tasks to the labelers according to their skills 
and the difficulty levels of querying tasks. 
 This becomes particularly challenging in streaming data scenarios due to future uncertainty, e.g., how to allocate the correction effort across the learning horizon. 

Online learning from streaming data corresponds to today's data common practise to train machine learning models with a small set of data that is periodically curated from the public domains. 
Machine learning models thus need to be learned online from the stream data. Moreover, due to privacy or regulation constraints~\cite{gdpr}, or storage limits, data turns are available for the limited duration.
Moreover, in such scenarios, combating label noise adds to the challenge. 

The prior art tries to enhance the robustness of the deep model training against noisy labels in the off-line scenario by (i) filtering out noisy data through model disagreement~\cite{han:NIPs18:coteaching,Yu:ICML19:Coteaching+}, (ii) correcting noisy labels through the estimated noisy corruption matrix~\cite{patrini2017making}, or (iii) modifying the loss functions~\cite{Ma:2018:ICML:D2L,wang:ICCV19:scl}. Among the related studies, 
high quality labels from human experts are used to correct labels through the estimate noise corruption matrix, e.g., the trust data set in Distillation~\cite{Li:ICCV17:Distilling} and GLC~\cite{hendrycks:NIPs18:glc}. The trust data sheds light on how a small percentage of high quality labels can prevent deep models from  accuracy degradation due to label noise. However, the focus on the off-line datasets renders existing approaches insufficient for handling noisy data streams only a subset of which can be learnt at a time. Indeed, on-line learning from noisy labels can lead to a much more severe degradation of accuracy than the typical off-line case. 
Moreover, such a subset of trust data is randomly selected and the cost of acquiring additional data label yet to be modeled. 





In this paper we address the challenge of training a deep classifier from noisy labeled data streams that are collected over time and that can only be learnt for a limited time. We propose an active learning framework, \alg, which aims to learn a robust classifier by cost-effectively assigning the label
correction task to either a strong or a weak labeler within a labeling budget. Extensively querying the strong labeler can easily lead to budget exhaustion, whereas the weak labeler might require multiple queries to achieve the cleansing goal. \alg consist of four steps: filtering out the suspicious data, choosing diverse samples, informativeness ranking, and selecting labelers. While training the classifier, \alg leverages the output of its classifier 
to filter suspicious data samples. As for label cleansing, we propose a labeler selection function Q that combines 
 the overall model confidence 
and the informativeness of individual samples. 

Our contributions are the following.
First we design a cost and skill aware active learning framework for noisy data streams. Secondly, by leveraging the 
diversity of the labelers we are able to greatly enhance the robustness of deep models even in challenging on-line learning scenarios. The proposed Q function can effectively assess the model confidence and informativeness of data samples 
and assign the suitable labelers accordingly. 





\section{Related Work}
\label{sec:related}
We summarize the related work in the context of robust learning, active learning, and streaming data---the main themes discussed in this paper.

\textbf{Robust learning} against noise labels. Training a robust deep classifier against noisy labels is an active research field~\cite{patrini2017making, Ma:2018:ICML:D2L, Reed:2015:ICLR:bootstrap, han:NIPs18:coteaching}. The main focus is on the off-line scenario where the training data is available at once and can be learned unlimited. To build robust classifier, the prior art either filters out the suspicious noisy data by the intermediate results of deep model~\cite{han:NIPs18:coteaching, Yu:ICML19:Coteaching+} or derives noise resilient loss function \cite{Ma:2018:ICML:D2L}, e.g., symmetric cross entropy~\cite{wang:ICCV19:scl}.
Konstantinov et al.~\cite{Konstantinov:ICML19:Untrusted:dirtylabel} considers multiple entrusted sources and assign weights to the sources according to their quality quantified by the difference between the source and target distributions.

\textbf{Active learning} 
Active learning \cite{settles2009active} is typically drawn to leverage the expert knowledge to provide the label information for the useful unlabeled data samples that might be limited in the real world data set \cite{Dy:TPMI18:probalisticActive, Chaudhuri:ICML18:ALloggedData}. Having multiple expert, as the experts may vary in their experience, it is important to match the expertise of labelers to the labeling tasks. While \cite{Chaudhuri:NIPs16:ImperfectLablers} considers imperfect labelers that may abstain from labeling, the study in~\cite{Huang:2017:IJCAI:DiverseLabelers} assumes having multiple labelers with different costs and qualities and actively selects both samples and the labelers considering sample usefulness and labeler's accuracy and cost, assuming that all the labelers are prone to make mistakes. The study~\cite{Sheng:2008:SIGKDD:Multilabeler} focuses on the selection of informative samples in the presence of several non-expert labelers via majority voting of the of the labelers. 
Considering the same framework, an extension of unbalanced labels is studied in~\cite{Zhang:2015:TransCybernetics:UnbalancedMultiLabeler}. However, they fail to leverage the labelers based on their expertise in labeling. 
In contrast,
\cite{Tanno:2019:CVPR:MultiAnnotatorConfusion} considers several noisy annotators with unknown expertise and jointly estimates the confusion matrix of the annotators and the true label distribution. 
Furthermore, the study \cite{Chaudhuri:NeurIPs15:StrongWeakLablers5} designs a difference classifier that predicts where the non-expert labeler differs from the expert in a binary classification task.
Moreover, a line of research focuses on multiple uncertain experts' view in clustering \cite{Dy:AISTATS18:partitionLaels, Dy:ICML17:undertainExperts}, however, they don't benefit from active learning for informative sample identification.


\textbf{Streaming Data} As an emerging challenge in real-world scenarios, online deep learning is largely overlooked by the prior art. Some studies have attempted to overcome the convergence problem of online deep learning using the idea of sliding window~\cite{Zhou:2012:AISTATS:onlineIncremental} and~\cite{Lee:2016:IJCAI:LifelongOnline}. the authors in \cite{SahooPLH18IJCAI} had a different approach to the problem by dynamically changing the depth of the network from shallow to deep to adapt the model capacity. Noisy data streams have been studied by~\cite{chu2004adaptive} and~\cite{zhu2006effective} using an ensemble of classifiers to build a robust model that maximizes the data likelihood. However, the focus of this research was on traditional machine learning models and is not scalable 
to deep networks.

\section{Dealing with Weak and Strong Labelers}
\label{sec:problem_def}

\begin{figure}[h]
	\includegraphics[width=0.9\textwidth, trim={0cm 0.5cm 0cm 0.8cm},clip]{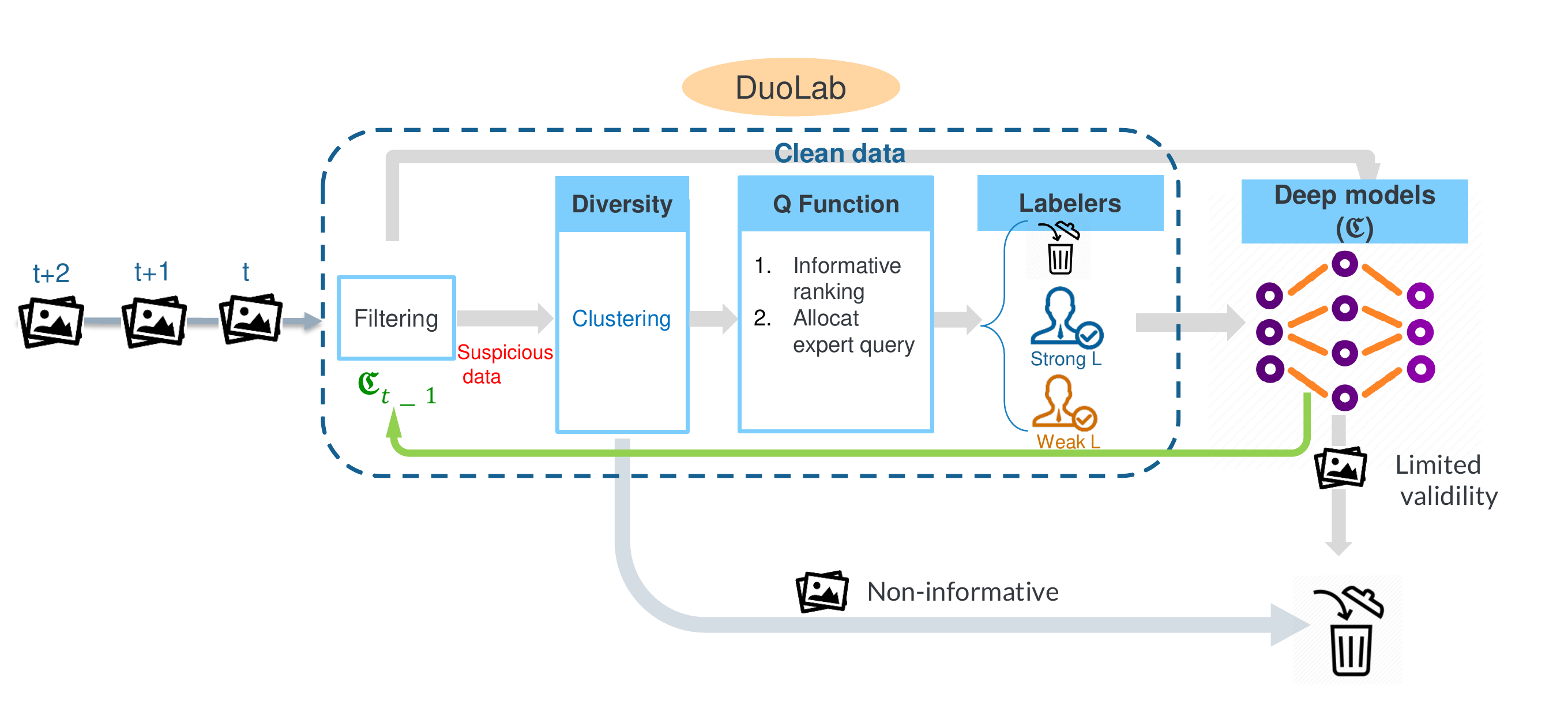}
	\centering
	\caption{On-line learning scenario with noisy labels: data streams, classifier, and strong/weak labelers  }
	\label{fig:arch}
\end{figure}

In this paper we consider the following on-line learning scenario, which is illustrated in Figure~\ref{fig:arch}. The data periodically, at consecutive time steps $t, t+1, \ldots$, streams into the classifier $\mathfrak{C}$ in small 
batches $D$ for training. 
Each arriving data batch is used for training the classifier for a limited duration. Here, we assume the most stringent case in which each batch of data can only be used for training the classifier 
until the arrival of the next batch. 
 
 Each data instance $(\mathbf{x}_j, \tilde{y}_j)$ contains feature inputs $\boldsymbol{x}_j \in \mathcal{X} \subset \mathbb{R}^d$ and a 
 potentially noisy label $\tilde{y} \in \mathcal{Y} := \{1, ..., N\}$. 
 Due to annotation errors or even malicious attacks~\cite{patrini2017making}, the collected label ($\tilde{y}$) can be 
 corrupted from the original label ($y$). The classifier can either learn directly from the whole batch of dataset 
 or it can first selectively correct labels and then learn from the 
 combined. clean and cleansed data. We will now go through the components 
 of our method \alg in Figure~\ref{fig:arch}; filtering, clustering and the Q function are explained in Section~\ref{sec:method}.

There are two types of labelers, namely a weak and a strong one
, available to correct the noisy labels subject to a budget. 
The weak labeler $\mathcal{W}$ is not an expert in labeling and instead of assigning a label to each unlabeled data instance, it can answer 
questions about its label with "yes/no". When presented with the true label, $\mathcal{W}$ gives  the answer "yes", and "no" otherwise. The strong labeler $\mathcal{S}$ on the other hand, knows the true label of the data; however, it is very costly to use. We assume $\mathcal{S}$ to be $\mathfrak{c}$ times as 
expensive as $\mathcal{W}$ to label each data point. 
Therefore, $\mathfrak{c}$ times asking $\mathcal{W}$ with "yes/no" questions is equivalent to asking $\mathcal{S}$ for the true label once. 

Ultimately, the objective is to train a classifier in the presence of noisy labels considering these two labelers with a limited budget $B$ per batch 
for labeling to achieve a certain level of accuracy. Detecting noisy samples and deciding which data to ask from which labeler is the main challenge of this problem. Since there exists a limited resource, i.e., the labeling budget, it is crucial to spend it on cleansing the representative and useful noisy samples. Therefore, the identification of the informative samples is one of the goals to solve this problem. Moreover, another challenge is, when relying on the weak labeler, how many "yes/no" questions to ask, and if requiring the strong labeler, how to allocate its budget for the arriving batch 
of data over time. To summarize, we aim to address the following questions in this paper: (i) how to detect noisy samples, (ii) how to identify informative data samples, 
and (iii) how to allocate the cleansing budget across the different labelers. 



\section{\alg}
\label{sec:method}
We propose an algorithm called \alg, which detects the noisy labeled data and decides how to find their true label within the budget and to train a deep multi-class classifier  
over time.
To cost-effectively train an accurate classifier, \alg leverages the information provided by the labelers as well as the model's prediction. \alg consists of the four stages of \textit{filtering}, \textit{clustering}, \textit{ranking} (informative sample selection), and \textit{labeler selection}. We explain each of these steps below
. Here, we specifically consider convolution neural networks (CNN) as the underlying classifier. The general design of \alg can be applied on different types of classifier.

\subsection{Filtering: Identifying the Suspicious Data}
\label{sec: filtering}
The first step of \alg is to identify the suspicious data samples that might have corrupted labels. 
We aim to filter out the noisy samples 
by leveraging the model's prediction and the confidence. We assume that the CNN classifier is initially trained with a small clean labeled set $D^I$. After the initial training, data samples $\{(\mathbf{x}_j,\tilde{y}_j)\}$ arriving in batches are provided to the model to predict each sample's label. To filter out the noisy samples, for each sample $\mathbf{x}_j$ we compare the highest and second-highest predicted label\footnote{The ranks of the predicted labels are determined by the value of softmax output of CNN.} 
$\hat{y}^1_j$ and $\hat{y}^2_j$ with the given label $\tilde{y}_j$, defining the sequence of $n$ highest predicted labels for $\mathbf{x}_j$ by $\mathfrak{C}$ as $\Omega=\{\hat{y}^n_j\}^N_{n=1}$. 
We consider the samples that have $\hat{y}^1_j = \tilde{y}_j$ or $\hat{y}^2_j= \tilde{y}_j$ to be clean, and add them to the clean set $C=\{(\mathbf{x}^c_j,y_j)\}$. The other samples are suspected to be noisy, and are collected in the suspicious set $U=\{(\mathbf{x}^u_j,\tilde{y}_j)\}$. 
Relaxing the filtering criteria to include lower-ranked predicted labels, e.g., the third and fourth-highest predicted labels, may lead to more false positives, i.e., noisy samples that are identified as clean ones. Contrarily, more stringent filtering criteria, e.g., only using the highest predicted label, results into more false negatives. 

\subsection{Clustering: Adding Diversity}
Since cleaning all the noisy samples is very expensive, our goal is to select samples that properly represent the 
 dataset and are highly informative. It has been shown~\cite{Wang:2017:TransFuzzySyst:Diversity} that only relying on the informative sample selection may
result in selecting similar samples that would cause a waste of the labeling budget. To add diversity to the selected samples, we apply kmeans clustering on the extracted features of the suspicious set $U$. The features are extracted from the network's last layer's output before the softmax. Next, we select the $\mathfrak{k}$ most informative samples of each cluster using active learning as explained in the next section, and discard the rest.

\subsection{Ranking: Informativeness}
When dealing with a large set of data that requires (re)labeling with a limited budget, the key is to identify the data samples the model could gain from the most when being trained on them. Active learning aims to find the most informative data and asks their true labels from the labelers. We rank the samples based on their informativeness, select the top $\mathfrak{k}$ ones from each cluster, and put them in the set $K=\{(\mathbf{x}^k_j,\tilde{y}_j)\}$. The samples in $K$ are then processed in decreasing order of their informativeness index $i$, where $i=1$ is the index of the most informative sample in $K$. 

To determine the informativeness of a sample, we estimate the prediction uncertainty of the network via the prediction probability $\mathbf{p}(\mathbf{x}_j)$  
at the output of the softmax layer. 
We employ a well-known active learning method called \textbf{Best-versus-second-best (BvSB)}~\cite{joshi:2009:CVPR:multi} to measure the uncertainty of the model while classifying a sample. 
This method considers samples for which the difference between the probabilities of the two most likely classes is the smallest, meaning that these samples could have easily been classified either way. 
That is, using $p_{best}$ and $p_{second-best}$ to denote the probabilities of the most likely and second most likely class,
the smaller the value of $ I(\mathbf{x}_j)=p_{best}(\mathbf{x}_j)-p_{second-best}(\mathbf{x}_j)$,
the higher the informativeness of $\mathbf{x}_j$. 

\subsection{Labeler Selection: Cost Optimization} 
The next step is to decide which labeler to choose considering their cost to cleanse the noisy labels. 
We introduce a cost sensitive labeler selection function $Q$ that solicits the appropriate labeler for each data sample in $K$. We consider three criteria to be crucial in the process of labeler selection: \textit{i)} the informativeness of the sample, \textit{ii)} the cost of the labelers, and \textit{iii)} the reliability of the model . 
Our aim is to spend more budget on the informative samples, and therefore we set the value of the selection function to be higher if a sample is more informative, and to guarantee cleansing those samples we aim to assign the most informative samples to the strong labeler. 
As the second criterion is the cost of the labelers, the challenge is to decide how to allocate the limited budget between the strong and weak labelers.  We define the selection function to be lower if the cost is high, resulting in higher chance of the selection of the weak labeler in costly situations. 
Moreover, the higher the value of the function, the higher the chance to ask the strong labeler for labeling.
Furthermore, if the model is not performing well enough, it means we need more cleansing so the value of the function should increase. 


\textbf{Cost Sensitive Labeler Selection Function}.
We will now define the cost sensitive labeler selection function $Q$ and show how it 
takes into account the three labeler-selection criteria mentioned above.
Before we do so, we define the following measure $L_V(t)$ for the overall model reliability up to the current batch $t$
using the cross-entropy loss on the clean validation set $D^V$:
\begin{equation}
 \label{eq_reliability}
  L_V(t) = \sum_{v=1}^{V}p(\mathbf{x}_v,y)\log(p(\mathbf{x}_v,\hat{y}^1)). 
 \end{equation}
A lower value of $L_V(t)$ indicates a higher reliability. 
In addition, we denote by $E_\mathcal{W}(t)$ and $E_\mathcal{S}(t)$ the numbers of queries from the current batch $t$ that have been asked sofar from the two labelers. The function $Q$ is only applied when the current total cost does not exceed the budget, that is, when $E_\mathcal{W}(t) + \mathfrak{c}E_{}\mathcal{S}(t) \leq B$.  
The value of the labeler selection function $Q$ on a data sample $\mathbf{x}^u_j$ is now defined as
\begin{equation}{
\label{eq_cost_sel}
{Q(\mathbf{x}^u_j)= \frac{L_V(t)}{I(\mathbf{x}^u_j)\mathfrak{c}E_\mathcal{S}(t)},
}}
\end{equation}
which has a higher value when the model is less reliable, when the sample is more informative, and when the strong labeler $\mathcal{S}$ has not yet been queried very often---the more $\mathcal{S}$ is queried during a batch, the more difficult it
becomes to query it for later samples. 

Algorithm~\ref{alg:training3} shows the steps of our proposed method.
As indicated in the algorithm, for each data arrival, after detecting the noisy samples we calculate their values of the cost sensitive selection function $Q$. 
If $Q(\mathbf{x}^u_j) > \Bar{Q}$ for some threshold $\Bar{Q}$, $\mathbf{x}^u_j$ is presented to $\mathcal{S}$ to be labeled and added to the clean set $C$, otherwise we query $\mathcal{W}$. If the answer is \textit{"No"}, we 
repeat the steps until we get the clean label or we exceed the budget.
In the latter case, we discard the sample. 

\begin{algorithm}
\SetAlgoLined
\small

\SetKwInOut{Input}{Input}\SetKwInOut{Output}{Output}
\Input{Initial dataset $D^I$, Data batches $D$, weak labeler $\mathcal{W}$, strong labeler $\mathcal{S}$, cost of strong labeler $\mathfrak{c}$, maximum number of weak queries per sample $\Bar{w}$, budget $B$, clustering parameter $\mathfrak{k}$}
\Output{Training set $C$ for the classifier $\mathfrak{C}$}
\vspace{0.1cm}
Train $\mathfrak{C}$ with $D^I$\\
\vspace{0.1cm}
\ForEach{arriving $D=\{(\mathbf{x}_j,\tilde{y}_j)\}$ and each $(\mathbf{x}_j,\tilde{y}_j)$}{


\eIf {($\hat{y}^1_j = \tilde{y}_j$) \textbf{or} ($\hat{y}^2_j= \tilde{y}_j$) }
{$C=\{(\mathbf{x}^c_j,y_j),\mathbf{x}^c_j$ :=$\mathbf{x}_j\}$  \hfill  \CommentSty{\#clean set}}
{
$U=\{(\mathbf{x}^u_j,\tilde{y}_j),\mathbf{x}^u_j$ :=$\mathbf{x}_j\}$  \hfill  \CommentSty{\#suspicious set} \\
}
}
Apply Kmeans clustering on $U$.\\ From each cluster select the most informative $\mathfrak{k}$ samples and add to $K=\{(\mathbf{x}^k_j,\tilde{y}_j)\}$. \hfill \CommentSty{\#Discard the rest} \\

\For {$\mathbf{x}_i$ in $K$ where $i$ is the informativeness index}{
$w=0$\\
\uIf{($Q>\Bar{Q}$) \textbf{and} ($E_W(t)+\mathfrak{c}E_S(t)+\mathfrak{c} \leq B$) \vspace{0.1cm}}{Query $\mathcal{S}$, update $E_\mathcal{S}$ and $Q$\\ 
Add $\mathbf{x}_i$ to $C$ 
}
\uElseIf{($E_\mathcal{W}(t)+\mathfrak{c}E_\mathcal{S}(t)+1 \leq B$) \textbf{and} ($w<\Bar{w}$) }
{Query $\mathcal{W}$ based on $\Omega$, update $E_\mathcal{W}$ and $Q$, $w=w+1$ 
\\
\eIf{(The answer is "Yes")}{Add $\mathbf{x}_i$ to $C$ }{Go to step 14 }
}
\Else{Discard $\mathbf{x}_i$}
{
}
}

\caption{Algorithm of \alg: filtering, clustering, labeler selection, and training}
\label{alg:training3}
\end{algorithm}


A key challenge about the weak labeler is how to proceed upon receiving \textit{"No"} for a sample. A naive approach is randomly choosing from the remaining labels and continuing until receiving \textit{"Yes"}. Here, we introduce a more intelligent way 
by asking $\mathcal{W}$ for the label of each sample in the order of $\Omega$, which is the sorted sequence of the labels based on the classifier's prediction probabilities (see Section~\ref{sec: filtering}); this presumably results in fewer $\mathcal{W}$ queries. 
Furthermore, we introduce a parameter $\Bar{w}$ indicating the maximum number of queries to the weak labeler per sample, which prevents wasting too much of the budget on a single sample and allows for more exploration.

\section{Experiments} \label{sec:evaluation}
We evaluate \alg on CIFAR-10 and CIFAR-100 whose data instances are corrupted with label noise and are on-line streamed to the classifier. We compare \alg with the state of the art in different, varying learning scenarios, i.e., availability of labelers, cost, and the size of the on-line batch.  We demonstrate how \alg improves the accuracy of the classifier using filtering and cleansing by the weak and strong labelers within the limited budget.


    
    

    
    


\subsection{Experimental Setup}
\label{exp setup}
\textbf{Training Network's Characteristics}.
As \alg classifier for CIFAR-10 and CIFAR-100, we use two Convolutional Neural Network (CNN) architectures defined in \cite{Ma:2018:ICML:D2L} with ReLU activation functions, softmax activation as image classifier and cross-entropy as loss function. We train the models by using stochastic gradient descent with momentum $0.9$, learning rate $0.1$, and weight decay $2\times10^{-3}$. 
\alg and all baselines are implemented using Keras v2.2.4 and Tensorflow v1.12, except co-teaching which uses PyTorch v1.1.0.

\textbf{Label Noise Injection} To inject noise into the labels, we select samples with the probability equal to the noise rate and replace the true label with a random different label with uniform probability. We use noise rates of $30\%$ and $60\%$ in the training set. The test and validation sets are clean. 

\textbf{Noise Resilient Baselines }
We put \alg in the context of other noise-resistant techniques drawn from the related work 
and adapted to the online scenario: \textbf{D2L}~\cite{wang2018iterative} estimates the dimensionality of subspaces during training to adapt the loss function, \textbf{Forward}~\cite{patrini2017making} corrects the loss function based on the noise transition matrix,
\textbf{Bootstrap}~\cite{Reed:2015:ICLR:bootstrap} (hard and soft) uses convex combination of the given and predicted labels for training 
, and
\textbf{Co-teaching}~\cite{han:NIPs18:coteaching} exchanges mini-batches between two networks trained in parallel.

\textbf{Training Parameters for CIFAR-10 and CIFAR-100}.
 For CIFAR-10, the initial training set consists of 4K samples, and 1K samples are considered as the validation set to measure model reliability. The remaining 55K samples are divided into 45K training and 10K testing set. The training data arrive in 45 batches of 1K samples with $30\%$ and $60\%$ noise in an online setting. The initial set is trained for 60 epochs and the rest are trained 20 epochs per batch. The epochs are selected in a way to let the network train enough but also avoid overfitting due to small batch size. 
 We set the cost sensitive function threshold $\Bar{Q}$ to $10$ and the cost of the strong labeler to $\mathfrak{c}=\{2,10\}$. 

For CIFAR-100 we increase the arriving data batch size to 9K and 60 epochs per batch to cope with the higher complexity. The cost sensitive function threshold $\Bar{Q}$ was set to $150$ and the cost of the strong labeler to $\mathfrak{c}=\{5,25\}$. In both datasets, the threshold is determined experimentally. In general, a high threshold, e.g., $\Bar{Q}=150$, leads to using less the strong labeler and more the weak labeler. 

For both CIFAR-10 and CIFAR-100, we set the labeler budget of \alg to be $25\%$ of the batch size, which amounts to $250$ and $2250$ unit cost per batch for each dataset, respectively. The higher budget leads to more cleansing and therefore, higher accuracy.  
Whenever the accuracy of the network over the validation sets drops for $r\%$ from the start to the end of a batch, we rollback the model weights to the previous batch's parameters before processing the following one. Rollback uses $r$ = 20\%. We report the highest accuracy among batches for each experiment. All the experiments are repeated three times and the average is reported in the tables. We show the highest accuracy among the first six rows of the tables \ref{labelers table cifar10} and \ref{labelers table cifar100} in bold.

\begin{table}[]
\caption{The accuracy, the 
average numbers of queries per labeler per batch, and the TP and FP (true and false positive) rates for \alg and different labeler selection baselines with budget $B=250$ and $\Bar{w}=2$ for CIFAR-10  ("-" means not applicable). 
}
\large
\renewcommand{\arraystretch}{1.1}
\resizebox{\linewidth}{!}{%
\begin{tabular}{l||crrrrr||rrrrr}
\hline
 \textbf{Noise} &   &   &  \multicolumn{1}{r}{\textbf{30\%}}  & &  & \textbf{} & \textbf{}  & & \multicolumn{1}{l}{\textbf{60\%}}&    &                 \\ \hline \hline

\multicolumn{1}{l||}{\textbf{Method}}       & \multicolumn{1}{c|}{\textbf{c}} & \multicolumn{1}{c}{\textbf{Acc(\%)}} & \multicolumn{1}{c}{\textbf{no. S}} & \multicolumn{1}{c}{\textbf{no. W}} & 
\multicolumn{1}{c}{\textbf{TP(\%)}} & \textbf{FP(\%)} & \multicolumn{1}{c}{\textbf{Acc(\%)}} & \multicolumn{1}{c}{\textbf{no. S}} & \multicolumn{1}{c}{\textbf{no. W}} & 
\multicolumn{1}{c}{\textbf{TP(\%)}} & \textbf{FP(\%)} \\ \Xhline{1pt}

\alg  & \multicolumn{1}{c|}{2}   & \textbf{76.13} & 22.0  & 57.0   
& 61.43 
& 3.81 
& \textbf{69.44}  & 38.4   & 63.0 
& 33.57 
& 7.84 
\\ \hline
\alg  & \multicolumn{1}{c|}{10}  & \textbf{75.45}    & 4.7   & 52.5   
& 60.90 
& 3.80 
& 67.42  & 9.8   & 51.8 
& 32.95
& 7.71
\\ \hline
\alg+ Kmeans & \multicolumn{1}{c|}{2}  & 75.99    & 15.3     & 67.1  
& 61.26
& 3.66
& 68.61 & 21.5 & 78.7 
& 33.38
& 7.84 
\\ \hline
\alg+ Kmeans & \multicolumn{1}{c|}{10} & 75.33  & 4.1 & 56.1   
& 60.70
& 3.78
& 67.53 & 21.5 & 78.7 
& 33.38
&7.84
\\ \Xhline{1pt}
Only $\mathcal{S}$   & \multicolumn{1}{c|}{10}  & 74.96  & 25.0   & -  
& 60.80
& 3.78
& 66.84    &    25.0 & - 
&  32.84
& 7.81
\\ \hline
Only $\mathcal{W}$   & \multicolumn{1}{c|}{-}    & 75.11   & -   & 67.6  
& 60.87
& 3.84
& \textbf{68.93}  & -   & 89.5 
& 33.53
& 7.75
\\ \Xhline{1pt}
Clean All Suspicious & \multicolumn{1}{c|}{-}   & 77.26 & 343.2     & -     
& 61.72
& 3.91 
& 75.71   
& - & 575.1 & 35.03
& 7.46
\\ \hline
No AL(only Filter)   & \multicolumn{1}{c|}{-}    & 73.60 & -   & -   
& 60.77
& 3.78
&  63.34 &    -&-  
& 32.26
& 8.11
\\ \hline

Opt Filter   & \multicolumn{1}{c|}{-}   & 77.78   & - & - 
& 70.00
& -  & 72.56 & - & - & 
40.00
& -   \\ \hline
No Filter    & \multicolumn{1}{c|}{10}  & 62.21  & 7.3 & 96.7  
& - & -  & 36.91   & 17.7   & 24.2   
& - & -       \\ \hline

\end{tabular}
}

\label{labelers table cifar10}
\end{table}

\subsection{CIFAR-10 Results}
\textbf{The effect of filtering}.
Here we compare our results with three baselines: \textit{Opt Filter}, where there is an optimal filter that can perfectly select all the clean samples and discard all the noisy ones, \textit{No Filter}, where there is no filtering, and \textit{No AL}, where we filter out samples using our filtering method but no further cleansing is done on the suspicious samples by active learning. As the results show in Table \ref{labelers table cifar10}, 
\textit{Opt Filter} has a high accuracy due to omitting all the noisy samples and training over a relatively large clean set (700 and 400 clean samples for $30\%$ and $60\%$ noise). However, despite having a smaller clean data and some noisy samples due to false positive (FP) detection, our method is able to get close in terms of accuracy. Comparing to the \textit{No AL} results, one can observe the effectiveness of our filtering together with active cleansing. 

\textbf{The effect of clustering}.
As Table \ref{labelers table cifar10} shows, clustering results in cleaning more suspicious samples compared to \alg without clustering; however, since it fails to cleanse fewer highly informative samples by $\mathcal{S}$, it results in lower accuracy.

\textbf{The effect of the labeler selection}. 
We compare \alg with two extreme labeler selection baselines: i) \textit{Only $\mathcal{S}$}, where only the strong labeler is available to be queried, and ii) \textit{Only $\mathcal{W}$}, where only the weak labeler is available to be queried. Moreover, to show the upper and the lower bound of the accuracy, we present the accuracy of the case where all the suspicious samples are cleansed and also when no cleansing is done, namely \textit{No AL}. As the results show in Table \ref{labelers table cifar10}, with lower cost for the strong labeler, i.e., with $\mathfrak{c}=2$, it is more beneficial to use only the strong labeler since it leads to cleaning more suspicious samples. The reason is that due to $\Bar{w}$, the weak labeler fails to give the true label if the true label is not among the top predictions of the classifier. However, with higher costs, using only $\mathcal{S}$ results in cleaning a very small number of samples. Contrarily, although using only the weak labeler might result in cleaning more data, missing some highly informative samples due to the limitation caused by $\Bar{w}$ results in a low performance. Our proposed method outperforms these two cases by benefiting from both cleansing the highly informative samples with the strong labeler, and minimizing the cost with the weak labeler. 

\subsection{CIFAR-100 Results}
\textbf{The effect of filtering}. 
As the results in Table \ref{labelers table cifar100} show, considering that CIFAR-100 has more classes than CIFAR-10 and less data per class, filtering is more difficult. 
However, we succeed in having a very low FP rate by our filtering method, and therefore we achieve an accuracy that is very close to the upper-bound of filtering \textit{Opt Filter}. Comparing the accuracy of \textit{No AL} and \alg with \textit{Opt Filter}, one could observe that filtering has a high impact on the performance, especially when the labelers' cleansed sample set is small (less than $2\%$ of the batch size), due to a low budget and more complicated dataset. Moreover, since filtering relies on the model's prediction, the higher accuracy of the model leads to a better filtering, i.e., a higher TP and a lower FP rate, which again impacts the accuracy itself.

\textbf{The effect of clustering}. Similar to CIFAR-10, clustering results in a slightly lower accuracy, since it affects the number of cleansed samples by the labeler. As shown in Table \ref{labelers table cifar100}, applying clustering on the suspicious data causes less cleansing by the strong labeler, which is equivalent to missing more informative samples.

\textbf{The effect of the labeler selection}. As Table\ref{labelers table cifar100} shows, \alg performs better than \textit{Only $\mathcal{S}$} and \textit{Only $\mathcal{W}$} in the case of high strong query cost. In this case, \alg succeeds in cleansing more samples than \textit{Only $\mathcal{S}$} and therefore achieves a similar or higher accuracy. This difference is more apparent with $60\%$ noise. 
Moreover, \alg achieves a similar cleansed number compared to \textit{Only $\mathcal{W}$} and due to benefiting form $\mathcal{S}$ beside $\mathcal{W}$, \alg guarantees the cleansing of the highly informative samples and thus results in a more accurate model. The higher the cost is, the more effective our method performs.

\begin{table}[]
\caption{The accuracy, the average numbers of queries per labeler per batch, and the TP and FP (true and false positive) rates for \alg and different labeler selection baselines with budget $B=2250$ and $\Bar{w}=2$ for CIFAR-100 ("-" means not applicable). 
}
\large
\renewcommand{\arraystretch}{1.1}
\resizebox{\linewidth}{!}{%
\begin{tabular}{l||crrrrr||rrrrr}
\hline
 \textbf{Noise} &   &   &  \multicolumn{1}{r}{\textbf{30\%}}  & &  & \textbf{} & \textbf{}  & & \multicolumn{1}{l}{\textbf{60\%}}&    &                 \\ \hline \hline

\multicolumn{1}{l||}{\textbf{Method}}       & \multicolumn{1}{c|}{\textbf{c}} & \multicolumn{1}{c}{\textbf{Acc(\%)}} & \multicolumn{1}{c}{\textbf{no. S}} & \multicolumn{1}{c}{\textbf{no. W}} & 
\multicolumn{1}{c}{\textbf{TP(\%)}} & \textbf{FP(\%)} & \multicolumn{1}{c}{\textbf{Acc(\%)}} & \multicolumn{1}{c}{\textbf{no. S}} & \multicolumn{1}{c}{\textbf{no. W}} & 
\multicolumn{1}{c}{\textbf{TP(\%)}} & \textbf{FP(\%)} \\ \Xhline{1pt}

\alg   & \multicolumn{1}{c|}{5}    & 39.07 & 33.3 & 102.1 
& 32.18
& 0.44 
& \textbf{34.45} & 44.0  & 171.9 
& 17.21
& 0.97
\\ \hline
\alg  & \multicolumn{1}{c|}{25}  & 39.27    & 7.3   & 102.7   
& 32.57 
& 0.45
& \textbf{34.14}  & 8.9   & 174.2 & 
17.65
&0.90
\\ \hline
\alg+ Kmeans & \multicolumn{1}{c|}{5}  & 38.45 & 16.9 & 105.1 
& 33.14
& 0.49
& 34.35 & 17.3 & 190.8 
&17.65
& 0.97
\\ \hline
\alg+ Kmeans & \multicolumn{1}{c|}{25}  & 38.98 & 4.0 & 100.5 
& 32.30
& 0.49 
& 32.92  & 4.0 & 181.4   
&17.11
&0.97
\\ \Xhline{1pt}
Only $\mathcal{S}$   & \multicolumn{1}{c|}{25} & \textbf{39.57} & 90.0 & - 
& 33.14
& 0.47
& 33.13 & 90.0 & - 
& 17.12
& 0.95
\\ \hline
Only $\mathcal{W}$   & \multicolumn{1}{c|}{-}   & 38.25   & -   & 105.9    
& 32.83
& 0.48
& 32.94 & -   & 192.2 
& 17.12
&0.99
\\ \Xhline{1pt}
Clean All Suspicious & \multicolumn{1}{c|}{-}   & 49.05  &   5085.4  & -   
&40.91
&0.45
& 49.53   & 6829.6   & - 
& 23.38
& 0.74
\\ \hline
No AL(only Filter)   & \multicolumn{1}{c|}{-}  & 38.08  &  - &  - 
& 32.63
&0.44
& 32.32 & -  & -  
& 16.18
& 0.98
\\ \hline
Opt Filter   & \multicolumn{1}{c|}{-}   &  42.28  & - & - 
& 70.00
& -  & 37.33 & - & - 
&40.00 
&-    \\ \hline
No Filter    & \multicolumn{1}{c|}{25}  & 30.86   & 112.0   & -  
& - & -  &   13.19 & 112.0 & - 
& - & -   \\ \hline

\end{tabular}
}

\label{labelers table cifar100}
\end{table}

\begin{wraptable}{R}{0.55\textwidth}
\vspace{-0.7cm}
\hfill
\begin{minipage}{0.54\textwidth}
\small
\caption{The accuracy of \alg and the noise resilient baselines for CIFAR-10 and CIFAR-100 with $30\%$ noise. 
}

\begin{tabular}{c|l|c|c}
\cline{2-4}
 \multicolumn{1}{c}{} & \multirow{2}{*}{\textbf{Method}} & \multicolumn{2}{c}{\textbf{Accuracy (\%)}}\\
\cline{3-4}
\multicolumn{1}{c}{} & & CIFAR-10 & CIFAR-100\\
\hline \hline
\multirow{5}{*}{\rotatebox[origin=c]{90}{\textbf{Baselines}}}
 & D2L             & $52.77$ & $11.55$\\ 
 & Forward         & $59.94$ & $25.56$ \\ 
 & Co-teaching     & $61.52$ & $29.41$\\
 & Bootstrap soft  & $48.95$ & $23.35$\\
 & Bootstrap hard  & $49.61$ & $24.02$ \\
\hline
\multirow{2}{*}{\rotatebox[origin=c]{90}{\textbf{Our}}}
 & \alg($\mathfrak{c}=2,5$)     & \textbf{$76.13$} & \textbf{$39.07$} \\
 & \alg($\mathfrak{c}=10,25$)     & \textbf{$75.45$} & \textbf{$39.27$} \\
\end{tabular}
\label{baselines}
\end{minipage}
\vspace{-0.3cm}
\end{wraptable}

\subsection{Comparison with the Noise Resilient Models}
We compare 
our proposed \alg with the baselines described in Section~\ref{exp setup} under $30\%$ label noise. Table~\ref{baselines} summarizes the results for different noise-resistant model baselines and \alg. Analyzing the results in the table shows that although these methods are well-known to combat the label noise, they achieve this by benefiting from a large dataset available for training, and therefore they fall short to adapt to the streaming data setting. On the contrary, \alg, benefiting from a novel filtering method and the knowledge of the experts, 
excellently adapts to the streaming data scenario with small batch size and for the limited learning duration.





\section{Conclusion}
In this paper we introduce \alg a method to overcome the challenge of training deep neural networks with streaming noisy labeled data by weak and strong labelers. As the strong labeler comes with its cost of labeling and is constrained by a a budget 
, our method decides when to use each labeler to achieve a high level of performance beside cleaning as more samples as possible.
As a result of \alg benefiting from the merits of both labelers in cleaning the highly informative samples and efficiently spending the budget, we are able to perform better than using only one of the labelers. We achieve this by only cleansing up to $10\%$ of the batch size in CIFAR-10 and $2\%$ in CIFAR-100.

\section*{Broader Impact}
This core idea of fusing human and artificial intelligence in \alg can be broadly applicable to data management and knowledge extraction. The cost- and privacy-awareness in \alg addresses the gap between state of the practise, i.e., streaming learning on noisy data, and the state of the art on off-line learning on the clean data. For instance, the rich information embedded in social media or recommendation systems is unfortunately noisy or even fake and  can not be easily extracted through the off-line resilient solutions. The proposed \alg has a great potential to distill the data and improve the knowledge extraction for the society.  On the other hand, 
if \alg fails to defend the training process against noise, the classifier can end up being inaccurate and may even harm the society in case of misuse.
The other downside of utilizing \alg is that we might indirectly foster the noisy data generation as the cost-efficiency of cleansing data via the labelers improves.

\medskip


\bibliographystyle{unsrtnat}
\bibliography{neurips_2020}
\end{document}